\documentclass[10pt,twocolumn,letterpaper]{article}

\usepackage{iccv}
\usepackage{times}
\usepackage{epsfig}
\usepackage{graphicx}
\usepackage{amsmath}
\usepackage{amssymb}

\usepackage{color}
\usepackage{ulem}
\usepackage{tabularx}
\usepackage{subcaption,float}
\usepackage{booktabs}
\usepackage{multirow}
\usepackage{amssymb}% http://ctan.org/pkg/amssymb
\usepackage{pifont}% http://ctan.org/pkg/pifont
\usepackage{wrapfig}
\newcommand{\cmark}{\ding{51}}%
\newcommand{\xmark}{\ding{55}}%

\usepackage[pagebackref=true,breaklinks=true,letterpaper=true,colorlinks,bookmarks=false]{hyperref}

\iccvfinalcopy % *** Uncomment this line for the final submission

\def\iccvPaperID{8611} % *** Enter the ICCV Paper ID here

\ificcvfinal\fi

\newif\ifcomments
\commentsfalse %Remove comments

\usepackage{enumitem}

\definecolor{red}{rgb}{.9,0.1,0.1}

\definecolor{green}{rgb}{0,0.65,0.35}

\begin{document}
%%%%%%%%% TITLE
% \title{M3D-VTON: Clothes-aware Controllable Monocular-to-3D Human Manipulation}
\title{M3D-VTON: A Monocular-to-3D Virtual Try-On Network} 

\author{Fuwei Zhao{$^{1}$},~ Zhenyu Xie{$^{1}$},~ Michael Kampffmeyer{$^{2}$},~ Haoye Dong{$^{1}$}\\\vspace{-16pt}\\ Songfang Han{$^{3}$},~ Tianxiang Zheng{$^{4}$},~ Tao Zhang{$^{4}$},~ Xiaodan Liang{$^{1*}$}\\\vspace{-10pt}\\
{$^{1}$}Shenzhen Campus of SYSU, {$^{2}$}UiT The Arctic University of Norway, {$^{3}$}UC San Diego, {$^{4}$}Momo \\
\small{\tt{\{zhaofw@mail2,xiezhy6@mail2,donghy7@mail2,xdliang328@mail\}.sysu.edu.cn}}\\ \small{\tt{michael.c.kampffmeyer@uit.no,s5han@eng.ucsd.edu,\{zhengtianxiang1128,allenxuejian\}@gmail.com}}
}


% \maketitle

\twocolumn[{%
\renewcommand\twocolumn[1][]{#1}%
\maketitle
    \includegraphics[width=1.0\hsize]{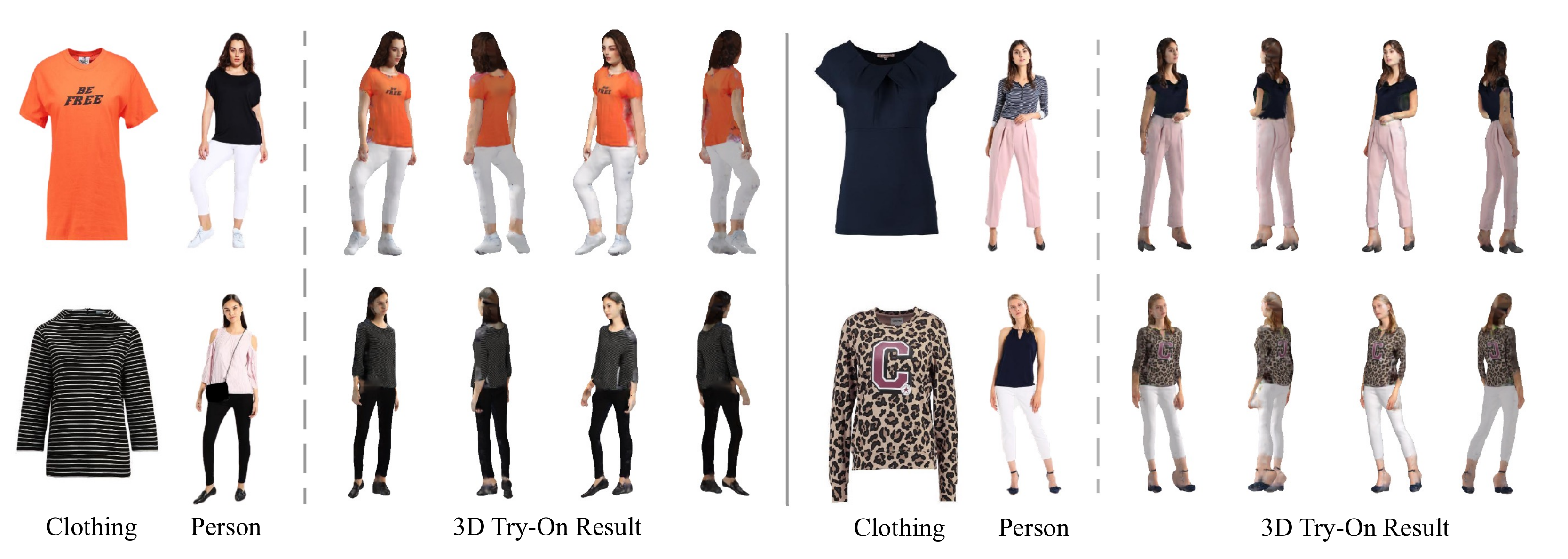}
    \vspace{-6mm}
    \captionof{figure}{Results of the proposed Monocular-to-3D virtual try-on network. Given the target clothing image and the reference person image, our M3D-VTON can reconstruct the 3D try-on mesh with the clothing changed and person identity retained.\newline\newline}
    \vspace{-4mm}
    \label{fig:teaser}
}]

\ificcvfinal\thispagestyle{empty}\fi

%%%%%%%%% ABSTRACT
\begin{abstract}
Virtual 3D try-on can provide an intuitive and realistic view for online shopping and has a huge potential commercial value.
However, existing 3D virtual try-on methods mainly rely on annotated 3D human shapes and garment templates, which hinders their applications in practical scenarios. 
2D virtual try-on approaches provide a faster alternative to manipulate clothed humans, but lack the rich and realistic 3D representation.
In this paper, we propose a novel Monocular-to-3D Virtual Try-On Network (M3D-VTON) that builds on the merits of both 2D and 3D approaches. 
By integrating 2D information efficiently and learning a mapping that lifts the 2D representation to 3D, we make the first attempt to reconstruct a 3D try-on mesh only taking the target clothing and a person image as inputs.
The proposed M3D-VTON includes three modules:
1) The Monocular Prediction Module (MPM) that estimates an initial full-body depth map and accomplishes 2D clothes-person alignment through a novel two-stage warping procedure; 
2) The Depth Refinement Module (DRM) that refines the initial body depth to produce more detailed pleat and face characteristics; 
3) The Texture Fusion Module (TFM) that fuses the warped clothing with the non-target body part to refine the results.
We also construct a high-quality synthesized Monocular-to-3D virtual try-on dataset, in which each person image is associated with a front and a back depth map. 
Extensive experiments demonstrate that the proposed M3D-VTON can manipulate and reconstruct the 3D human body wearing the given clothing with compelling details and is more efficient than other 3D approaches.~\footnote{code will be available at~\url{https://github.com/fyviezhao/M3D-VTON}}
\end{abstract}

%%%%%%%%% BODY TEXT
\section{Introduction}
3D virtual try-on, the process of fitting a specific clothing item onto a 3D human shape, 
%has attracted increasing attention due to its promising research value and commercial potential.
has attracted increasing attention due to its promising research and commercial value.
Recently, researchers' interest has moved from physics-based~\cite{Baraff98largesteps,10.1145/566654.566623, 10.5555/846276.846281,10.1109/TVCG.2008.79, peng2012drape,fabian2014scs} or scan-based approaches~\cite{gerard2017clothcap, lahner2018deepwrinkles, stoll2010video} to learning-based 3D try-on methods~\cite{bharat2019multigarment,chaitanya2020tailornet,aymen2020pixsurf,zhu2020deepfashion3d,corona2021smplicit}, dressing a 3D person directly from 2D images and getting rid of costly physics simulation or 3D sensors. 
However, most of these learning methods~\cite{bharat2019multigarment,chaitanya2020tailornet,aymen2020pixsurf} build on the parametric SMPL~\cite{matthew2015smpl} model and depend on some predefined digital wardrobe~\cite{bharat2019multigarment}, limiting their real-world applicability.
Moreover, the inference speed of these existing 3D approaches is still insufficient, largely due to the optimization cost introduced by the parametric 3D representation.

Related to this, research on image-based virtual try-on aims to fit an in-shop clothing onto the target person and has been explored intensively~\cite{han2018viton,wang2018toward,yu2019vtnfp,han2019clothflow,yang2020acgpn,thibaut2020swuton,dong2019mgvton}. Most of these works utilize the Thin Plate Spline (TPS) transformation~\cite{bookstein1989TPS} to achieve the clothes-person alignment and fusion, obtaining photo-realistic try-on results. These 2D methods are attractive due to their small computation cost and extensive amount of available training data on shopping websites. Nevertheless, their try-on results are in 2D image space and ignore the underlying 3D body information, leading to inferior capability of representing the human body.

To address the above limitation of 2D/3D approaches, we propose a light-weight yet effective Monocular-to-3D Virtual Try-On Network (M3D-VTON), which integrates both 2D image-based virtual try-on and 3D depth estimation to reconstruct the final 3D try-on mesh.
M3D-VTON consists of three modules as shown in Fig.~\ref{framework}. 
The first part is the Monocular Prediction Module (MPM), which utilizes a single network to serve the following three purposes: 1) regressing the parameters for the TPS~\cite{bookstein1989TPS} transformation; 2) predicting the conditional person segmentation that is compatible with the in-shop clothing; 3) estimating the full-body depth map.
Different from the warping operation in existing 2D try-on methods, MPM first utilizes a novel self-adaptive affine transformation to transform the in-shop clothing to the appropriate size and location before the non-rigid TPS deformation. 
The second part is the Depth Refinement Module (DRM), which jointly uses the estimated depth map, the warped clothing, the non-target body part and the image gradient information to enhance the geometric details in the depth map. In particular, DRM introduces a depth gradient loss to better exploit the high-frequency details in the inputs.
Finally, the Texture Fusion Module (TFM) leverages the 2D information (e.g., warped clothing) and the 3D information (e.g., estimated full-body depth) to synthesize the try-on texture.
The collaborative use of the 2D information and the body depth map provides instructive information for the synthesizing process. 
Given the estimated 2D try-on texture and the refined body depth map, M3D-VTON obtains a colored point cloud and reconstructs the final textured 3D virtual try-on mesh.

We conduct extensive experiments on the new MPV-3D dataset, which is constructed by running PIFuHD~\cite{shunsuke2020pifuhd} on the existing MPV dataset~\cite{dong2019mgvton}.
Compared with other 3D try-on methods, M3D-VTON recovers detailed body shapes and realistic texture color while being more computationally efficient.
Our main contributions are:
\begin{itemize}[itemsep=1pt,topsep=1pt]
\item We are the first to exploit the merits of both 2D and 3D approaches to solve the monocular-to-3D try-on problem. Our approach reconstructs realistic 3D clothed humans while being faster than pure 3D methods. 

\item To facilitate more accurate geometric matching between the clothes and the reference person image, we introduce a self-adaptive pre-alignment strategy.

\item We utilize the available shadow information in the images and incorporate a novel depth gradient constraint to guide the network to capture and recover intricate geometric changes.

\item We construct a new synthesized 3D virtual try-on dataset, MPV-3D, which may stimulate the development of the Monocular-to-3D virtual try-on field. Extensive experiments show the surprising shape recovery and texture generation ability of our M3D-VTON.
\end{itemize}

\section{Related Work}

{\bf 2D Virtual Try-on.} 2D virtual try-on aims to transfer a target clothing onto a reference person. A series of works~\cite{han2018viton,wang2018toward,yu2019vtnfp,yang2020acgpn,dong2019mgvton,neuberger2020oviton,thibaut2020swuton,hsieh2019fashionon} have utilized the non-rigid TPS transformation~\cite{bookstein1989TPS} to obtain appealing virtual try-on results. 
Most of these works build upon VITON~\cite{han2018viton}, which proposes a coarse-to-fine architecture that first warps the in-shop clothing by TPS and then renders the final try-on result.
CP-VTON~\cite{wang2018toward} further trains a geometric matching module and uses a composition mask to better fuse the clothes and person. VTNFP~\cite{yu2019vtnfp} utilizes body segmentation as the synthesis guidance, producing clearer skin texture. ACGPN~\cite{yang2020acgpn} proposes a second-order constraint on TPS parameters to stabilize the warping process. 
Our method not only inherits the benefits of the aforementioned methods but also generates realistic 3D clothed human, providing an economic solution for monocular-to-3D virtual try-on.
%\medskip

\begin{table}
    \centering
    \def\arraystretch{0.75}
    \setlength{\tabcolsep}{5pt}
    %\begin{tabular}{m{5cm}}
    \begin{tabular}{lcccccc}
    \bf{Methods} & \bf{CC} & \bf{3D} & \bf{FBT} & \bf{SG} & \bf{ED} & \bf{FI} \\
    \midrule
    VITON~\cite{han2018viton} & Y & N & N & N & Y & Y \\
    CP-VTON~\cite{wang2018toward} & Y & N & N & N & Y & Y \\
    ACGPN~\cite{yang2020acgpn} & Y & N & N & Y & Y & Y \\
    \midrule
    PIFuHD~\cite{shunsuke2020pifuhd} & N & Y & N & N & N & Y \\
    MGN~\cite{bharat2019multigarment} & Y &Y & Y & Y & N & N \\
    DeepFashion3D~\cite{zhu2020deepfashion3d} & N & Y & N & Y & N & - \\
    Pix2Surf~\cite{aymen2020pixsurf} & Y & Y & N & N & Y & Y \\
    \midrule
    Deephuman~\cite{tang2019deephuman} & N & Y & N & Y & N & Y \\
    FACSMILE~\cite{Smith_2019_ICCV} & N & N & Y & N & Y & N \\
    NormalGAN~\cite{wang2020normalgan} & N & Y & Y & N & N & Y \\
    \midrule
    M3D-VTON(ours) & Y & Y & Y & Y & Y & Y \\
    \bottomrule
    \end{tabular}\vspace{-3mm}
    \caption{Comparison of M3D-VTON to related work in terms of their properties with Yes (Y) or No (N). The first three rows are 2D try-on methods, the middle rows are 3D try-on/reconstruction methods, and the bottom rows except ours are human depth estimation methods. Categorized based on: Changeable Clothes (CC); Clothed 3D Body (3D); Full Body Texture (FBT); Semantic Guidance (SG); Easy-to-get Dataset (ED); Fast Inference (FI).}
    \vspace{-8mm}
    \label{tab:methodscompare}
\end{table}

{\bf 3D Virtual Try-on.} Compared to the tasks of 3D human reconstruction and performance capturing~\cite{Zheng_2018_ECCV, Habermann_2020_CVPR,Gilbert_2018_ECCV,Saini_2019_ICCV,Yu_2019_CVPR,kanazawaHMR18,SMPL-X:2019,kocabas2019vibe,Huang_2020_CVPR,alldieck2019tex2shape}, 3D virtual try-on is more challenging due to the complex deformation of clothes.
PIFuHD~\cite{shunsuke2020pifuhd} provides a high-fidelity single-view textureless 3D human reconstruction pipeline that produces realistic clothing details, however, it can not perform garment transfer.
MGN~\cite{bharat2019multigarment} can predict parametric garment geometry and layer it on top of the SMPL~\cite{matthew2015smpl} model. Thanks to the layered representation, MGN can dress varying body shapes and poses but is limited to garments from their predefined digital wardrobe. 
DeepFashion3D~\cite{zhu2020deepfashion3d} provides more 3D clothes data to achieve more challenging clothing reconstruction. Pix2Surf~\cite{9156907} also aims to transfer more in-the-wild clothes images onto the SMPL model by learning dense correspondences between 2D garment silhouettes and UV maps of 3D garment surfaces.  However, both DeepFashion3D and Pix2Surf can not show the body texture. Besides, almost all these methods require a scanned 3D dataset for training, which is expensive to collect compared with our proposed high-quality synthesized dataset. Our method can recover both clothed body shape and texture, providing a more practical solution for 3D try-on.
%\medskip

{\bf Human Depth Estimation.} Recently, non-parametric 3D human reconstruction has been proposed to better capture shape details by predicting depth maps. Moulding Humans~\cite{Gabeur_2019_ICCV} estimates the front and back depth map from a single RGB image to generate a textureless 3D human. FACSMILE~\cite{Smith_2019_ICCV} is similar and adds a normal constraint to carve local depth details but manipulates naked bodies and is not cloth-aware. DeepHuman~\cite{Tang_2019_ICCV} also utilizes a normal map to refine the estimated depth but only generates the frontal body part, limiting its practical application. NormalGAN~\cite{wang2020normalgan} further uses an adversarial learning framework conditioned on normal maps to recover the textured 3D human body. However, NormalGAN requires the ground-truth depth map as input, which needs to be collected using expensive depth sensors. Compared with the above methods, our M3D-VTON is trained on high-quality synthesised data and allows for cloth-aware human manipulation. For ease of comparison, Table~\ref{tab:methodscompare} presents an overview of the properties of M3D-VTON and the most related approaches.

\section{M3D-VTON}
\begin{figure*}
\begin{center}
\includegraphics[width=0.9\hsize]{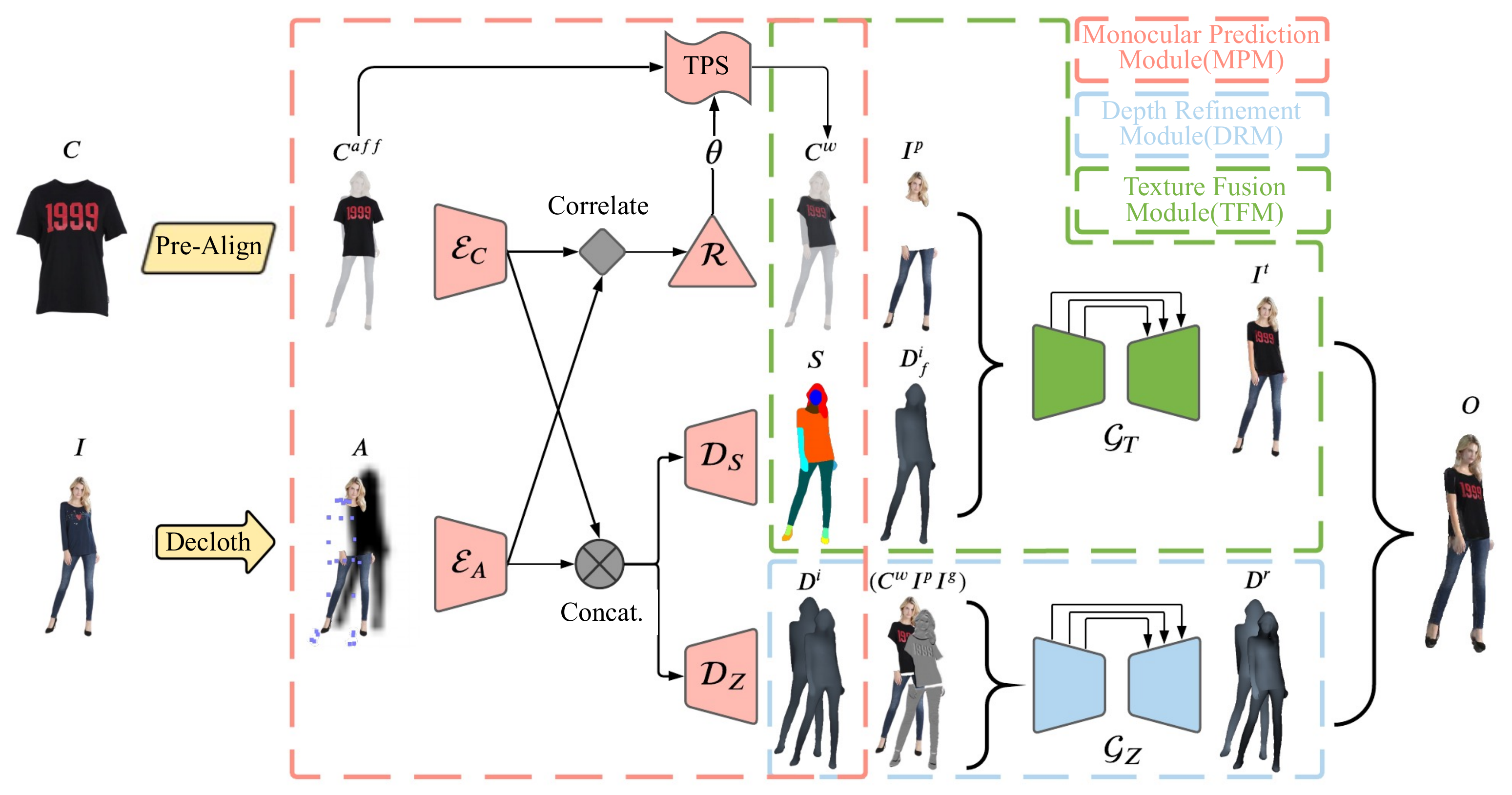}
\end{center}
    \vspace{-6mm}
   \caption{
Overview of the proposed M3D-VTON. The pipeline contains three modules with the following tasks.
a) \textbf{Monocular Prediction Module (MPM)}: obtaining the cloth-agnostic person representation $A$ through the decloth process, deforming the in-shop clothing $C$ to the warped clothing $C^w$ via a self-adaptive pre-alignment followed by a TPS transformation, predicting a person segmentation $S$, and estimating an initial double-depth map $D^{i}$.
b) \textbf{Depth Refinement Module (DRM)}: given the double-depth map $D^{i}$, the warped clothing $C^{w}$, the preserved person part $I^{p}$, and their shadow information $I^{g}$ as inputs, this module refines the initial depth map and produces more local details (like cloth folds and face structure) by incorporating a novel depth gradient constraint.
c) \textbf{Texture Fusion Module (TFM)}: rendering the results $I^{t}$ under the guidance of the semantic layout from MPM by fusing the warped clothes and the preserved texture information. 
Once $I^{t}$ and the refined depth map $D^{r}$ are spatially aligned, forming an RGB-D representation, we can directly get colored point clouds and triangulate them to obtain the 3D clothed human $O$ wearing the target clothes and with its identity preserved.
   }
  \vspace{-6mm}
\label{framework}
\end{figure*}

To facilitate 3D virtual try-on, we propose a novel Monocular-to-3D Virtual Try-On Network (M3D-VTON) that takes a clothing image $C$ and a person image $I$ as inputs, and reconstructs a 3D try-on mesh $O$ with clothes changed and person identity preserved. 
As illustrated in Fig.~\ref{framework}, M3D-VTON is composed of the Monocular Prediction Module (MPM), the Depth Refinement Module (DFM), and the Texture Fusion Module (TFM). 

%-------------------------------------------------------------------------

%-------------------------------------------------------------------------

\subsection{Monocular Prediction Module}
This module plays a preparatory role in the proposed M3D-VTON. It provides constructive guidance for the other two modules by warping the in-shop clothing, predicting a conditional person segmentation, and by estimating a base 3D shape using a multi-target network. All these tasks can be accomplished by utilizing the features extracted from the target clothing $C$ and the clothing-agnostic person representation $A$. $A$ consists of a 25-channel pose map (obtained by applying OpenPose~\cite{openpose} on person image $I$), a 3-channel unchanged person part ($I^p$) (obtained by applying \cite{liang2018look} on $I$), and a 1-channel coarse person mask that have been concatenated. We explain the three sub-branches of MPM in the following sections.

\textbf{Clothing Warping Branch.} 
Inspired by~\cite{rocco17cnngeometric}, the first branch of the MPM utilizes an end-to-end trainable geometric matching network to achieve the texture-preserving clothing-person alignment. Specifically, as part of the geometric matching network, the features extracted by the encoders $\mathcal{E}_C$ and $\mathcal{E}_A$ are fed into the feature correlation layer to calculate the matching score, which is used by the regressor $\mathcal{R}$ to predict the TPS transformation~\cite{bookstein1989TPS} parameters $\theta$ (see Fig.~\ref{framework}). 
However, directly estimating $\theta$ is non-trivial since there is a huge gap in size between the in-shop clothing $C$ and the arm-torso region of the reference person $I^{at}$. We therefore extract $I^{at}$ from $I$ by applying person segmentation~\cite{liang2018look} and design a \textbf{self-adaptive pre-alignment} procedure to transform $C$ to the proper position and size before conducting the TPS transformation. We formulate the procedure as an affine transformation:
\begin{equation}
C^{aff}=\left[\begin{array}{ll}
R & 0 \\
0 & R
\end{array}\right]C+\left[\begin{array}{l}
x^{c}_{I^{at}}-x^{c}_{C} \\
y^{c}_{I^{at}}-y^{c}_{C}
\end{array}\right],
\label{affine}
\end{equation}
where $C^{aff}$ denotes the transformed clothing item (see Fig.~\ref{framework}), $\left(x^{c}_{I^{at}},y^{c}_{I^{at}}\right)$ and $\left(x^{c}_{C},y^{c}_{C}\right)$ represent the center of $I^{at}$ and $C$, respectively. $R$ is a rescaling factor computed by comparing the aspect ratio to ensure that the aligned clothing is larger than or at least equal to the arm-torso region:
\begin{equation}
R=\left\{\begin{array}{ll}
\frac{h^{at}_{I}}{h_{C}}, & \frac{w_{C}}{h_{C}} \geqslant \frac{w^{at}_{I}}{h^{at}_{I}} \\
\frac{w^{at}_{I}}{w_{C}},&  \frac{w_{C}}{h_{C}}<\frac{w^{at}_{I}}{h^{at}_{I}}.
\end{array}\right.
\end{equation}

\begin{figure}
  \centering
  \includegraphics[width=1.0\hsize]{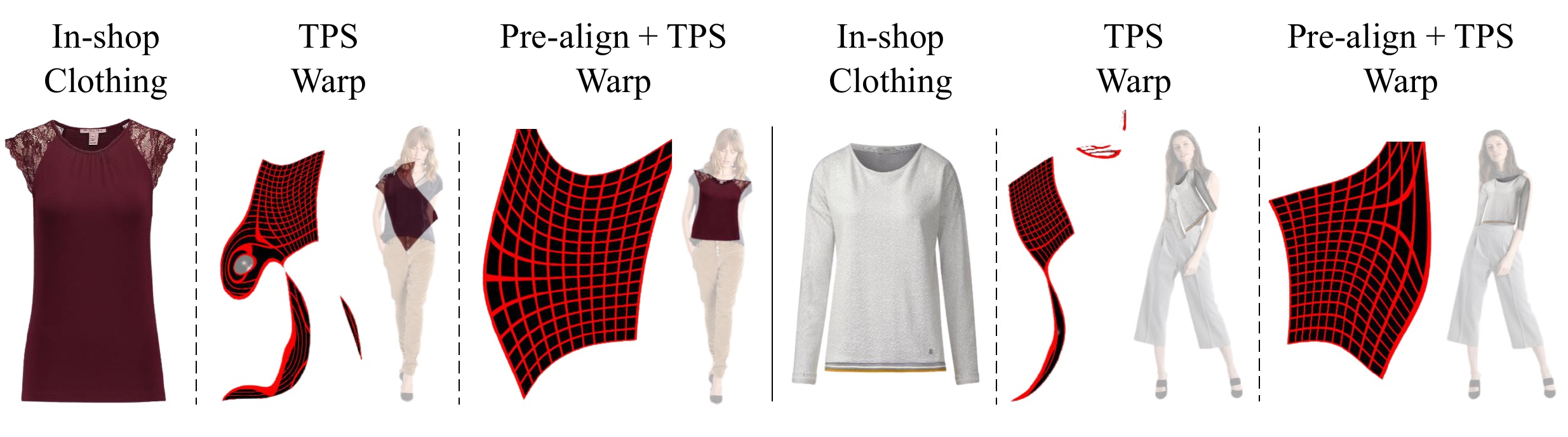}
  \vspace{-6mm}
  \caption{Verification of the pre-alignment strategy. The proposed self-adaptive transformations increase the quality of the warping (as shown in the third column).} 
  \vspace{-4mm}
  \label{warpcompare}
\end{figure}

%An intuitive understanding of Eq.~\ref{affine} is to first center align $C$ and $I^{at}$, following by thumbnailing $C$ to roughly the same size as $I^{at}$ to relieve pressure of the TPS warping. 
An intuitive understanding of Eq.~\ref{affine} is that it first center aligns $C$ and $I^{at}$ and scales $C$ to roughly the same size as $I^{at}$ to simplify the TPS warping step. The effectiveness of the alignment procedure is illustrated in Fig.~\ref{warpcompare}. 

Given $C^{aff}$, we pass both $C^{aff}$ and the clothing-agnostic person representation $A$ to the geometric matching network to regress the TPS parameters $\theta$, which are then used to warp $C^{aff}$ to the warped clothing $C^w$. 
During training, the difference between $C^{w}$ and the ground-truth $I^{c}$ (clothing-on-person) is used to define the warping loss:
\begin{equation}
\mathcal{L}_{w}=\|C^{w}-I^{c}\|_1.
\end{equation}

\textbf{Conditional Segmentation Estimation Branch.} 
The goal of this branch is to estimate the person segmentation supposing now wearing the desired clothing, which delineates different parts of the reference person (e.g., the sleeve-arm boundary). The segmentation mask provides inpainting guidance for the following texture fusion module to mitigate skin texture degradation or clothing-skin penetration especially for the case of self-occlusion or large clothing variation. 
As shown in Fig.~\ref{framework}, the feature maps from $\mathcal{E}_C$ and $\mathcal{E}_A$ are concatenated together and sent to the segmentation decoder $\mathcal{D}_S$ to generate the conditional person segmentation $S$. Although only \emph{paired} $\left(C, I\right)$ images\footnote{Reference person $I$ is wearing clothing $C$.} are fed to the model during training, the network can generalize to unpaired data at inference time due to the benefit of its clothes-agnostic representation. During training, we use the pixel-level cross-entropy~\cite{gong2017lip} $\mathcal{L}_{s}$ to optimize this branch.
\medskip

\textbf{Depth Estimation Branch.} 
%\commentM{Did you conduct an ablation study that shows the benefit of including the clothing branch to predict $D_Z$?} \responseF{Currently not, but I'll conduct this experiment. Although I think the clothing feature can help delineate cuff and skin when predicting depth, it is not supported by ablation study. }
The last branch in MPM aims at estimating a base 3D shape of the reference person. We represent the 3D shape in a double-depth form similar to~\cite{Gabeur_2019_ICCV}, i.e. a front and a back depth map corresponding to the respective sides of the 3D human representation. In this branch, the concatenated feature map is upsampled by the depth decoder $\mathcal{D}_{Z}$ to generate the front and the back depth.
During training, the loss function can be formulated as:
\begin{equation}
L_{z}=\| D_{f}^{i}-D_{f}^{gt}\|_1  + \| D_{b}^{i}-D_{b}^{gt}\|_1,
\end{equation}
where $D_{f}^{i}$ and $D_{b}^{i}$ represent the estimated front and back depth, and the superscript $i$ means ``initial''. $D_{f}^{gt}$ and $D_{b}^{gt}$ are the corresponding ground-truth depth maps. 

We refer to the estimated depth maps as ``initial'' depth since there are not enough clues for $\mathcal{D}_{Z}$ to infer the complete details of the warped clothing, such as the pleat details.
To obtain more precise 3D information, the initial depth map will be refined in the depth refinement module, which will be explained in Section~\ref{DRM}.

We train the three branches together within a multi-target network and combine the three aforementioned losses to yield the full loss of MPM:
\begin{equation}
    \mathcal{L}_{\text{MPM}} = \mathcal{L}_w + \mathcal{L}_s + \mathcal{L}_z.
\end{equation}

\subsection{Depth Refinement Module}
\label{DRM}
The reasons that the initially estimated depth map from MPM fails to capture geometric details (e.g., clothing details, face characteristics) are twofold: (1) the inputs of the MPM lack the warped clothing, which is crucial to carve clothing pleats; 
(2) the L1 depth loss used in MPM tends to penalize low-frequency differences between the estimated and the ground truth depth map, resulting in an over-smoothed depth result. \quad To add high-frequency depth details, we propose the Depth Refinement Module (DRM), which further exploits the brightness changes in the warped clothing $C^{w}$ and the preserved person part $I^{p}$ to refine the initial depth map. 
Specifically, we apply the Sobel operator on $C^{w}$ and $I^{p}$ and concatenate the gradient images to obtain the image gradient $I^g$, representing the changes in brightness.
Then, $I^g$, $C^{w}$, $I^{p}$ and the initial depth map $D^i$ are sent to an UNet-like generator $\mathcal{G}_{Z}$ to produce the refined depth map $D^r$.
During training, we propose two special losses to enable the network to capture the high-frequency details. Firstly, inspired by~\cite{hu2018revisiting}, we replace the vanilla L1 depth loss with a Log-L1 version, which penalizes close points more heavily and therefore guides the estimation to focus on intricate local details, which is formulated as:
\begin{equation}
    \mathcal{L}_{\text{depth}}=\frac{1}{n} \sum_{i=1}^{n} \ln\left(\epsilon_{i}+1\right),
\end{equation}
where $\epsilon_{i}$ is the L1 loss of the i-th depth point, and n is the total number of the front/back depth map points.

Secondly, to further strengthen the depth estimation and capture geometric details especially at the boundary of adjacent body parts, we incorporate a \textbf{depth gradient loss} :
\begin{equation}
    \mathcal{L}_{\text{grad}}=\frac{1}{n} \sum_{i=1}^{n} \left(\ln\left(\nabla_{x}\left(\epsilon_{i}\right)+1\right) + \ln\left(\nabla_{y}\left(\epsilon_{i}\right)+1\right)\right), \label{eq:gradloss}
\end{equation}
where $\nabla$ denotes the Sobel operator. 

Note that normal maps can be generated from depth gradient maps~\cite{7335535} and that Eq.~\ref{eq:gradloss} thus also penalizes the difference in normal maps. It is shown in~\cite{wang2020normalgan} that normal maps tend to contain more detailed geometric information than depth maps, therefore constraints along the normal direction can help recover geometric details and delineate the boundary of adjacent body parts, where the depth gradient is generally large.

The above two losses work in a complementary manner to constrain different types of errors:
a) $\mathcal{L}_{\text{depth}}$ ensures consistency along the z-direction, b) $\mathcal{L}_{\text{grad}}$ does the same for the x-, y- and thus normal direction. We therefore utilize a weighted sum of the aforementioned losses to train DRM:
\begin{equation}
\mathcal{L}_{\text{DRM}} = \lambda_{\text{depth}}\mathcal{L}_{\text{depth}} + \lambda_{\text{grad}}\mathcal{L}_{\text{grad}},
\end{equation}
where $\lambda_{\text{depth}}$, $\lambda_{\text{grad}}$ are set to 1.0, 0.5 respectively.

\subsection{Texture Fusion Module}
To synthesize photo-realistic body texture for the final 3D human mesh, we propose the Texture Fusion Module (TFM) which fuses the warped clothing with the unchanged person part to render seamless try-on results. 
TFM takes the preserved person part $I^{p}$, the warped clothing $C^w$, the predicted segmentation $S$, and the estimated initial front depth $D^{i}_{f}$ as inputs, and generates a coarse try-on result $\Tilde{I^c}$ as well as a fusion mask $\Tilde{M}$. 
The 2D clues of $I^{p}$, $C^w$, and $S$ provide the person appearance, clothing texture, and semantic guidance for the network. %Apart from the 2D clues,
% which are used in most of the existing 2D try-on approaches~\cite{han2018viton,wang2018toward,yu2019vtnfp,yang2020acgpn}, 
Further, TFM also considers the body depth map $D^{i}_{f}$, which contains the spatial information of different body parts along the z-axis. Under the extra guidance of $D^{i}_{f}$, TFM is capable of synthesizing the try-on result more precisely even in challenging self-occlusion cases.
Finally, the fusion mask $\Tilde{M}$ is used to fuse $C^w$ and $\Tilde{I^c}$ into the refined try-on result $I^{t}$, which can be formulated as:
\begin{equation}
    I^t=C^w\odot\Tilde{M}+\Tilde{I^c}\odot (1-\Tilde{M}).
\end{equation}

TFM is trained using the perceptual loss $\mathcal{L}_{\text{perc}}$~\cite{johnson2016perceptual} between the refined try-on result $I^t$ and the real person image $I$, the L1 loss $\mathcal{L}_{\text{try-on}}$ between $I^t$ and $I$, as well as the L1 loss $\mathcal{L}_{\text{mask}}$ between the estimated fusion mask $\Tilde{M}$ and the real clothing-on-person mask $M$. The combined loss for TFM can thus be formulated as:
\begin{equation}
    \mathcal{L}_{\text{TFM}} = \mathcal{L}_{\text{perc}} + \mathcal{L}_{\text{try-on}} + \mathcal{L}_{\text{mask}}.
\end{equation}

In the end, we can unproject the front-view and the back-view depth maps from DRM to get the 3D point clouds and triangulate them with screened Poisson reconstruction~\cite{10.1145/2487228.2487237}. Since the try-on result from TFM is spatially aligned with the depth map, it can directly be used to color the front side of the mesh. As for the back texture, we first inpaint the try-on image using the fast matching method proposed in~\cite{journals/jgtools/Telea04}, filling the face area with the surrounding hair color, and then mirror the inpainted ``back'' view image to texture the backside of the mesh.
This allows us to successfully achieve the monocular-to-3D conversion, producing the reconstructed 3D clothed human with retained identity. 
%retains the identity of the reference person and achieves the purpose of virtual try-on.

\begin{figure*}[t!]
  \centering
  \includegraphics[width=1.0\hsize]{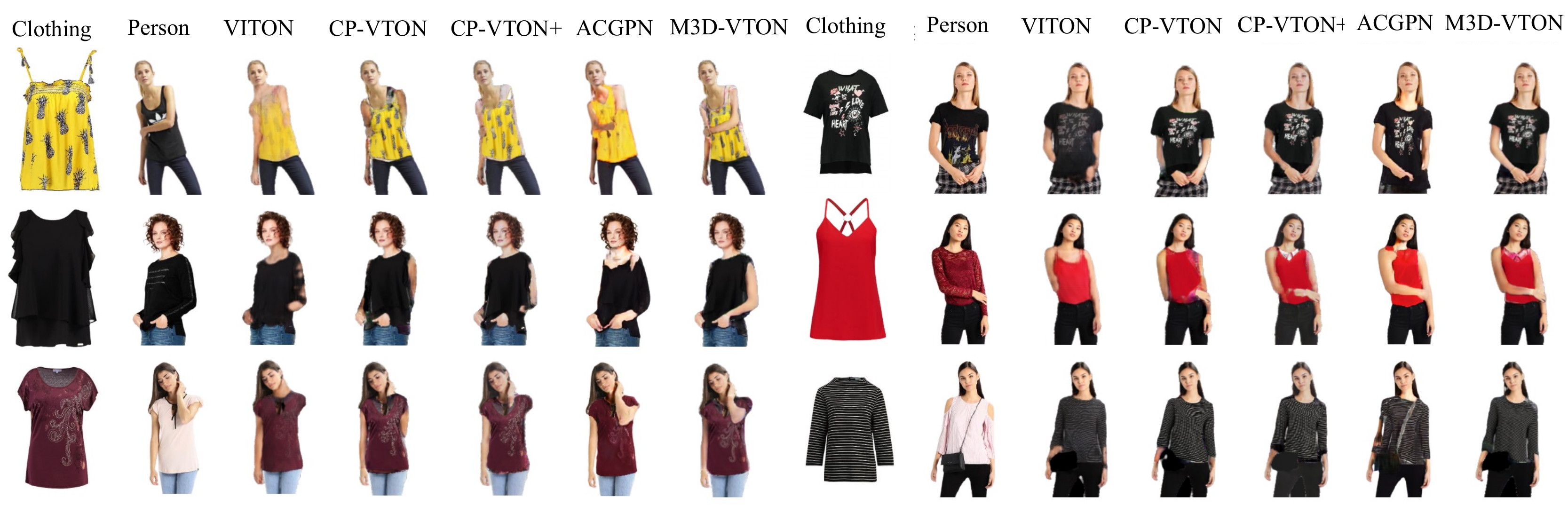}
   \vspace{-6mm}
  \caption{Qualitative comparison for the 2D try-on task. The first columns represent the inputs, columns 3 to 6 are prior approaches, and column 7 illustrates our proposed approach. 
  %\updateF{ACGPN added. Need to rewrite caption of this figure.} \commentM{Great job on adding ACGPN!} \responseF{Does this figure look too small?}\commentM{I think it is still ok, but it should not get smaller.} 
  %\commentM{Qualitative comparison for the 2D try-on task. The first two columns represent the inputs, columns 3 to 6 are prior approaches, and column 7 illustrates our proposed approach.} \responseF{Thanks for updating caption!}
  }
  %Visual comparison with the other methods in the texture fusion module. The first two columns are the inputs, the 3rd, 4th, and 5th columns are prior approaches, and the 6th to 9th columns illustrate different settings of this module (w/o depth and parsing, w/ only parsing, w/ only depth and w/ both depth and parsing guidance). The proposed M3D-VTON produces more realistic results due to its adaptive fusion.}
 \vspace{-3mm}
  \label{fig:2d_results}
\end{figure*}

\section{Experiments}
% We first describe the dataset used in our experiments and the implementation details, and then provide qualitative and quantitative comparisons for both the 2D and 3D try-on settings. We also analyze the effectiveness of different modules of our proposed M3D-VTON in the ablation study.

% \begin{wrapfigure}{r}{4.5cm}
% \centering
% % \flushleft
% % \flushright
% \includegraphics[width=1.1\hsize]{LaTeX/figures/mpv3d.png}
% \vspace{-6mm}
% \caption{Dataset Generation Process.}
% \vspace{-4mm}
%   \label{fig:mpv3d}
% \end{wrapfigure}

\subsection{Dataset Generation}
We construct the first monocular-to-3D try-on dataset MPV-3D based on the MPV dataset~\cite{dong2019mgvton}, which contains person images covering a wide range of poses and upper-body garments\footnote{Examples are shown in the supplementary.}. 
% \xd{Some examples are shown in Figure XX}\commentM{Should we still have another figure with some examples in the supplementary? I would argue that it can't hurt \footnote{Examples are shown in Figure XX in the supplementary.}}.\xzy{add examples in supplementary}
%For the first time, we construct a new dataset MPV-3D based on the MPV dataset~\cite{dong2019mgvton} for the monocular-to-3D try-on task.
MPV-3D contains 6566 clothes-person image pairs $(C,I)$ of size 512×320, in which each person image is associated with a front and a back depth map, $D_{f}$ and $D_{b}$, respectively. We obtain the depth maps and set them as the pseudo ground truth of our M3D-VTON by applying PIFuHD~\cite{shunsuke2020pifuhd} on the full-body front-faced person images from the MPV dataset and then orthographically projecting the generated human mesh to the double-depth maps. 
The dataset is further divided into a train set and a test set with 5632 and 934 four-tuples $\left(C,I,D_{f},D_{b}\right)$ respectively, and the test set is shuffled to form the unpaired $\left(C,I\right)$ list for quality evaluation.

\subsection{Implementation Details}
%We train the MPM first and train the DRM and TFM in parallel, since the MPM provides the inputs for the other two modules.
% \xd{missing details of network design?}
The MPM is trained separately from the DRM and the TFM as it provides the inputs to these modules, while DRM and TFM are trained together\footnote{We provide the complete architecture details in the supplementary.}.
%We train the MPM first and train the DRM and TFM in parallel, since the MPM provides the inputs for the other two modules.
Each module is trained for 100 epochs using the Adam optimizer, with $\beta_{1}=0.5$, $\beta_{2}=0.999$, and the learning rate is initialized as 0.0002 with a linearly decay to 0 in the last 50 epochs. The batch size is 8. The model is implemented in Pytorch and trained on a single NVIDIA 2080ti GPU.
During training, the reference person wears the same clothing as the target in-shop clothes as the try-on result for unpaired clothes and persons are not available as supervision. %can not beforehand have the try-on result of unpaired clothes and person as supervision. 
However, during testing, the target clothing is different from the clothing on the person and inference is performed in an end-to-end manner.

\begin{figure*}[t]
  \centering
  \includegraphics[width=1.0\hsize]{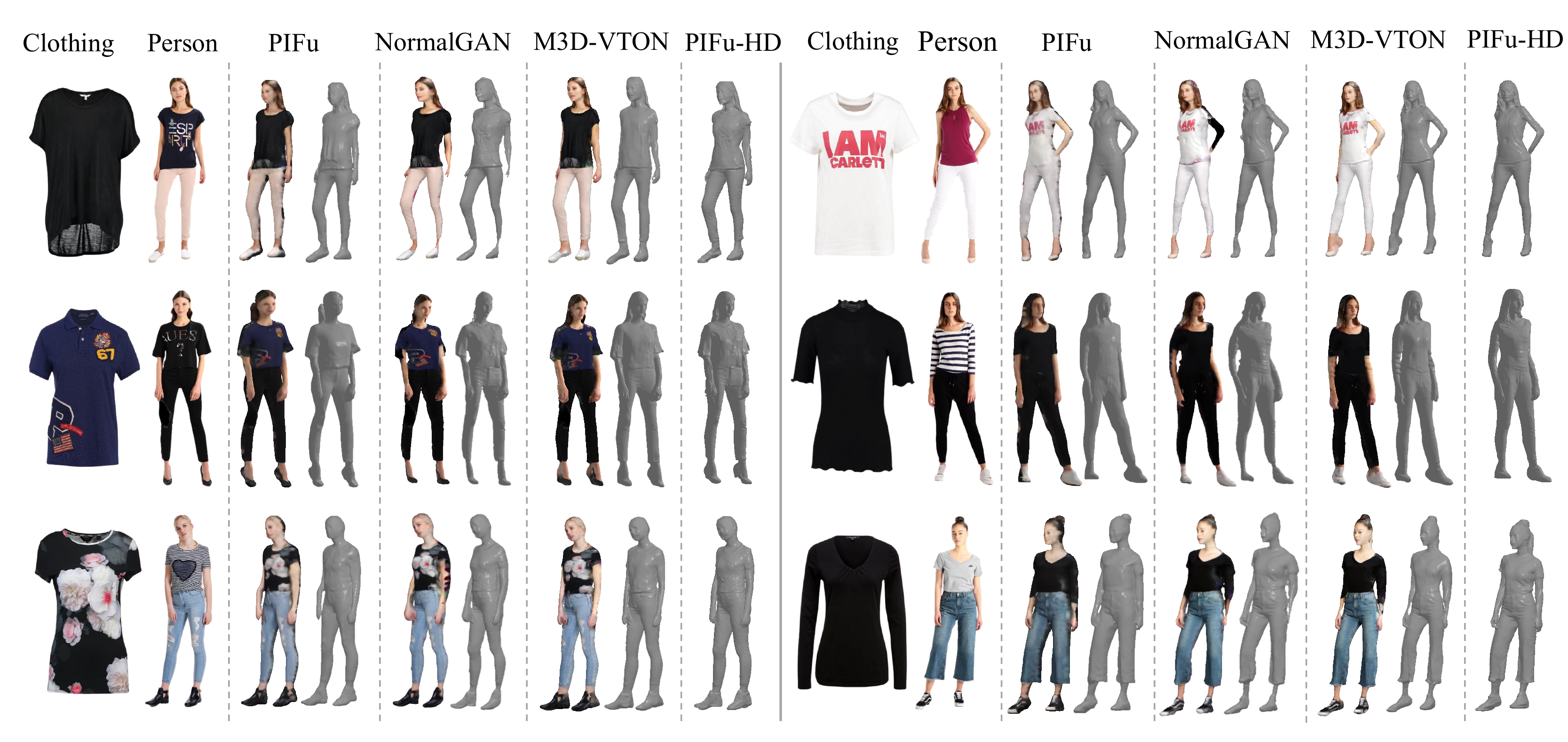} 
  \vspace{-6mm}
  \caption{Qualitative comparisons of 3D try-on results. The first two and the last columns respectively represent the inputs and the (pseudo) ground truth PIFu-HD mesh, while the others are 3D try-on results (w/ and w/o texture) from the different methods. The human mesh generated by our M3D-VTON contains more texture details and a more accurate shape compared to PIFu~\cite{saito2019pifu} and NormalGAN~\cite{wang2020normalgan} (note that NormalGAN uses GT front depth map as its input). 
  %The first two columns are the inputs, the others columns are the 3D clothed human results from different methods \commentM{Here it is the 3D try-on results and the human mesh, right?} \responseF{Yes. E.g. the 3rd column is the textureless 3D mesh, the 4th column is the textured 3D mesh, and so on.}
  }
  \vspace{-4mm}
  \label{fig:3d_results}
\end{figure*}

\begin{table}
    \def\arraystretch{0.75}
    \centering
    \begin{tabular}{cccc}
    \toprule
    Method & SSIM $\uparrow$ & FID $\downarrow$ & HE $\uparrow$ \\
    \midrule
    VITON~\cite{han2018viton} & 0.8807 & 28.43 & 21.35\% \\
    CP-VTON~\cite{wang2018toward} & 0.8503 & 20.05 & 10.65\% \\
    CP-VTON+~\cite{Matiur2020cpvton+} & 0.8782 & 23.18 & 12.57\% \\
    ACGPN~\cite{yang2020acgpn} & \textbf{0.8924} & 20.19 & 13.50\% \\
    \midrule
    M3D-VTON & 0.8804 & \textbf{20.04} & \textbf{41.92\%} \\
    \bottomrule
    \end{tabular}
    \caption{Quantitative comparisons to other 2D try-on methods. For fair comparison, we crop and resize our full-body try-on results to half-body like those in Fig.~\ref{fig:2d_results}, as other methods originally perform half-body try-on.
    \vspace{-5mm}
    }
    \label{tab:table1}
\end{table}

\subsection{2D Try-on Comparison with SOTA methods}
We compare our 2D try-on results with the existing state-of-the-art 2D try-on methods: VITON~\cite{han2018viton}, CP-VTON~\cite{wang2018toward}, CP-VTON+~\cite{Matiur2020cpvton+}, and ACGPN~\cite{yang2020acgpn}.

A qualitative comparison is shown in Fig.~\ref{fig:2d_results}. VITON lacks texture details of the clothing and fails to synthesize arms in self-occlusion cases. Although CP-VTON and CP-VTON+ can better preserve clothing texture, they perform poorly when the clothing is occluded by body parts. ACGPN fails to synthesize complete arms and may synthesize artifacts in the clothes region due to the stochasticity introduced by its segmentation estimation network.  Due to our two-stage warping strategy, M3D-VTON more accurately preserves the clothing texture, and synthesizes body parts precisely through the collaborative guidance of the conditional segmentation and the body depth map.

% The results of the quantitative comparison are illustrated in Table~\ref{tab:table1}.
For the quantitative comparison, we adopt the Structural SIMilarity index measure (SSIM)~\cite{Wang2004SSIM} and the Fr$\mathbf{\acute{e}}$chet Inception Distance (FID)\cite{heusel2017fid} to measure the similarity between the synthesized and the real images. Further, we conduct a human evaluation (HE) to assess the 2D try-on results from M3D-VTON and the other four baselines. Specifically, we invited 26 volunteers to complete a questionnaire that contains 40 assignments. In each assignment, given a person image and a clothing image, the volunteers are required to select the most realistic try-on image out of the ones produced by the five methods.

As shown in Table~\ref{tab:table1}, M3D-VTON obtains the lowest FID and highest human evaluation score, outperforming other baseline methods. Its SSIM score is on-par with the best performing model. 
To fairly compare with baseline methods which trained on the half-body VITON dataset~\cite{han2018viton}, the baseline methods take the cropped half-body images from MVP-3D as inputs and synthesize half-body results during testing. The full-body results of M3D-VTON are cropped to half-body images (as shown in Fig.~\ref{fig:2d_results}) following the same cropping procedure.

\subsection{3D Try-on Comparison with SOTA methods}
Since this is the first work that explores the monocular-to-3D virtual try-on setting, we design three hybrid models to conduct 3D try-on comparisons. Specifically, we first obtain the 2D virtual try-on result using CP-VTON and then generate the 3D try-on mesh using the state-of-the-art 3D human reconstruction approaches PIFu~\cite{saito2019pifu}, NormalGAN~\cite{wang2020normalgan}, and  Deephuman~\cite{tang2019deephuman}.
The qualitative and quantitative comparisons are shown in Fig.~\ref{fig:3d_results} and Table~\ref{tab:table2}, respectively. Since Deephuman does not recover the backside of the 3D shape, we compare with it only quantitatively. 

\begin{table}
\def\arraystretch{0.75}
    \centering
    \setlength{\tabcolsep}{1pt}
    \begin{tabular}{ccccc}
    \toprule
    Method & Abs. $\downarrow$ & Sq. $\downarrow$ & RMSE $\downarrow$ & HE $\uparrow$ \\
    \midrule
    Deephuman~\cite{tang2019deephuman} & 17.35 & 1.271 & 22.44 & - \\
    NormalGAN~\cite{wang2020normalgan} & 15.41 & 0.778 & 18.94 & 21.3\% \\
    PIFu~\cite{saito2019pifu} & 8.376 & 1.813 & 27.57 & 11.3\% \\
    \midrule
    M3D-VTON (ours) & \textbf{7.880} & \textbf{0.385} & \textbf{11.27} & \textbf{67.4\%}  \\
    \bottomrule
    \end{tabular}
    \vspace{-2mm}
    \caption{Quantitative comparisons to other 3d human reconstruction methods. All values have been divided by $10^{-3}$ for readability. Note for PIFu and M3D-VTON, we average their double-depth score. 
    The scores for Deephuman and NormalGAN are computed from single depth since they only recover front and back depth, respectively.}
    \vspace{-6mm}
    \label{tab:table2}
\end{table}

In Fig.~\ref{fig:3d_results}, the hybrid CP-VTON+PIFu model produces plausible 3D shape results but fails to recover detailed texture due to its unreliable implicit texture color inference. Unlike PIFu, NormalGAN uses the double-depth representation and directly sets the 2D image as mesh texture. However, NormalGAN requires a noisy ground truth depth map as input to infer the back shape and although we simulate the depth generation process in NormalGAN, it still tends to produce over-slim 3D persons. 
Compared with these hybrid methods, our M3D-VTON generates more realistic 3D persons and preserves detailed texture within a single model. 

The results of the quantitative comparison are shown in Table~\ref{tab:table2}. We use three common depth estimation metrics: Absolute Relative error (Abs.), Squared Relative error (Sq.) and Root Mean Squared Error (RMSE). Our method outperforms the benchmark models on all of the four measurements including human evaluation (HE), illustrating the superior shape generation ability of M3D-VTON. Finally, our method takes about 4 seconds to run for a given MPV-3D image pair (most computational cost occurs during the poisson reconstruction process), which is clearly faster and more efficient than pure 3D virtual try-on (such as Multi-Garment Net~\cite{bharat2019multigarment}, roughly 17s to run) or 3D human reconstruction (such as PIFu~\cite{saito2019pifu}, roughly 10s to run) methods.

% \usepackage{multirow}
\begin{table}[]
\def\arraystretch{0.75}
\centering
\begin{tabular}{c|cc|cc}
% \hline
\toprule
\multirow{2}{*}{MTM} & Pre-align           & IoU                 &Pre-align     & IoU \\ \cmidrule{2-5}
                     & \xmark       & 0.708 & \cmark              & \textbf{0.737}  \\
\midrule
\multirow{5}{*}{TFM} & Segmt. $S$  & Depth $D_f^i$  & SSIM $\uparrow$  & FID $\downarrow$ \\ \cmidrule{2-5}
                     & \xmark              & \xmark              & 0.9348       & 16.52   \\
                     & \cmark              & \xmark              & 0.9434       & 16.01 \\
                     & \xmark              & \cmark              & 0.9418             & 15.96   \\
                     & \cmark              & \cmark              & \textbf{0.9435}    & \textbf{15.74}   \\
\midrule
\multirow{4}{*}{DRM} & Grad. $I^g$        & Loss $\mathcal{L_{\text{grad}}}$       & Sq. $\downarrow$  & RMSE $\downarrow$ \\ \cmidrule{2-5}
                     & \xmark              & \xmark                                 & 0.0824  & 5.8369  \\    
                     & \cmark              & \xmark                                 & 0.0896  & 5.7650  \\
                     & \cmark              & \cmark                                 & \textbf{0.0801}  & \textbf{5.7420}  \\
\bottomrule
\end{tabular}
\vspace{-2mm}
\caption{The ablation study on the three modules. Note values of $\text{Sq.}$ and $\text{RMSE}$ are divided by $10^{-3}$ for readability. Here the scores of DRM are for front depth, as we do not have the back-side gradient image. And the TFM scores are computed from full-body try-on results.}
\vspace{-5mm}
\label{tab:ablation}
\end{table}

\subsection{Ablation Study}
We conduct ablation experiments on the three modules of M3D-VTON to verify their effectiveness.

\begin{figure}
  \centering
  \includegraphics[width=1.0\hsize]{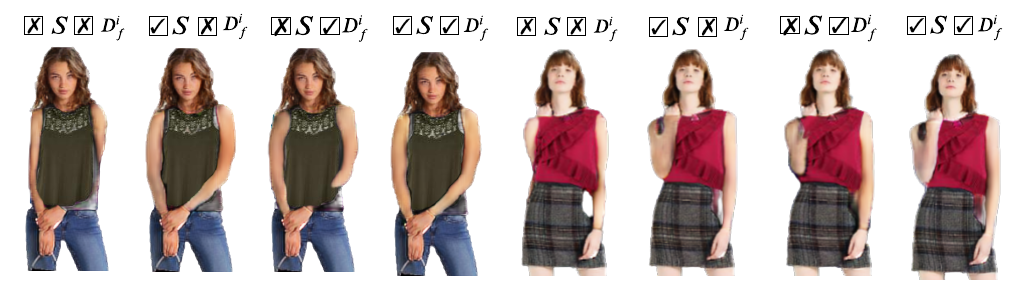}
  \vspace{-6mm}
  \caption{Visual comparisons to verify the effectiveness of the segmentation guidance and depth guidance in TFM.}
  \vspace{-3mm}
  \label{fig:tfmcompare}
\end{figure}

\begin{figure}
  \centering
  \includegraphics[width=0.95\hsize]{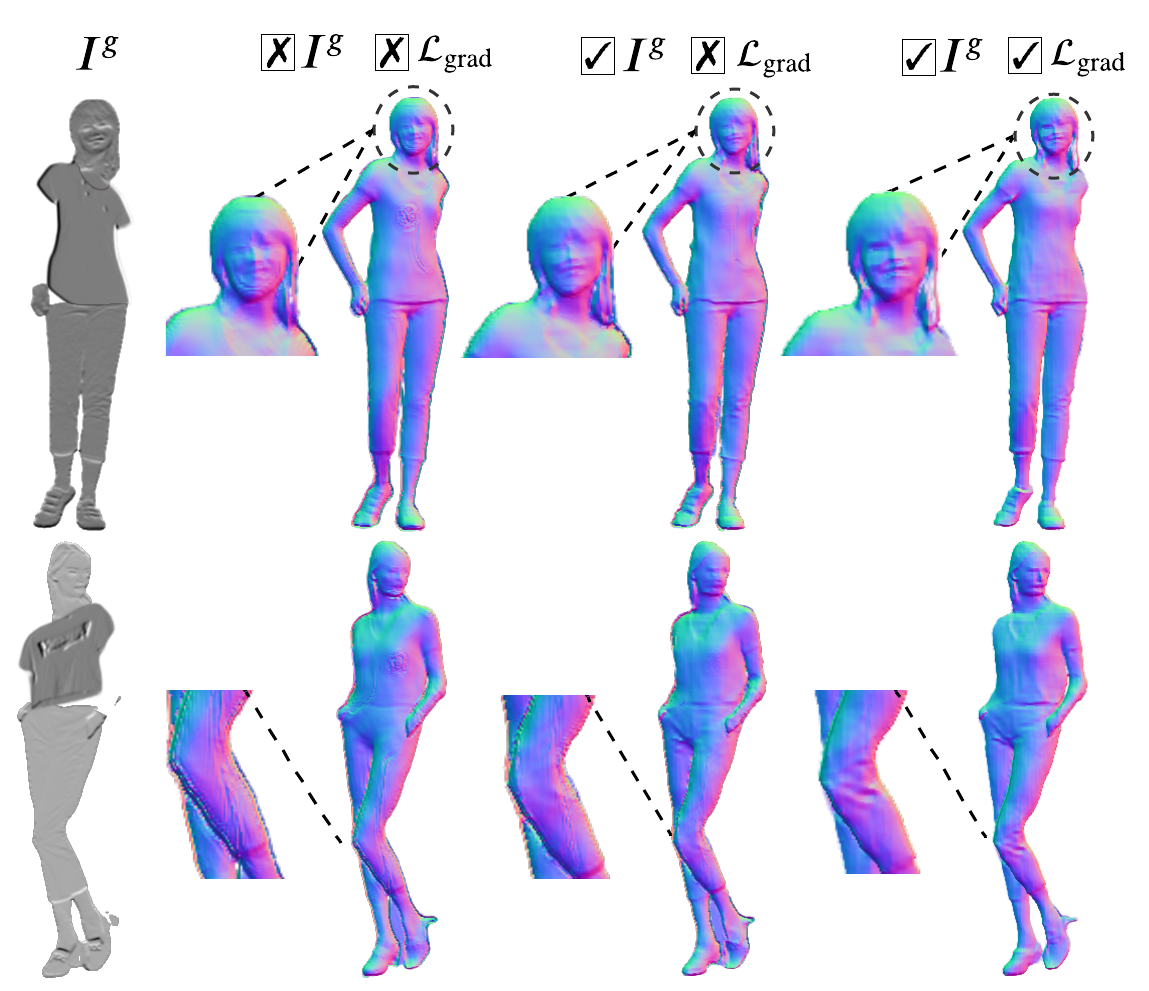}
  \vspace{-4mm}
  \caption{Visualized ablation study on DRM. $\mathcal{L}_{\text{grad}}$ is beneficial for producing geometric details (see last column).}
  \vspace{-5mm}
  \label{drmcompare}
\end{figure}

{\bf Effectiveness of the Self-Adaptive Pre-Alignment in MPM.} 
Fig.~\ref{warpcompare} shows that directly applying TPS results in excessive deformation and fails to warp clothes properly. Our two-stage warping with pre-alignment, instead, can generate gentle deformation and obtains precisely warped clothes. Quantitative results (Table~\ref{tab:ablation}, 1st row), verify this as the IoU between the warped clothes and the clothes region from the reference person increases with pre-alignment.

{\bf Effectiveness of Depth and Segmentation Guidance in TFM.} 
Fig.~\ref{fig:tfmcompare} illustrates the need that using these two guidance independently can help alleviate the self-occlusion issue. Furthermore, under their collaborative guidance, TFM can further improve the fidelity of arms in the synthesized results.
Table~\ref{tab:ablation} corroborates that they both contribute positively to the M3D-VTON. Note that the SSIM and FID scores here are calculated on the full-body results, while the scores in Table~\ref{tab:table1} are reported for the cropped half-body results to fit the setting of ACGPN for fair comparison. 

{\bf Effectiveness of depth gradient constraint in DRM.} Table~\ref{tab:ablation} and Fig.~\ref{drmcompare} show that the image gradient inputs and the proposed depth gradient constraint can improve the depth prediction and guide the DRM to carve more intricate details onto the 3D shape. The black dotted circles in Fig.~\ref{drmcompare} highlight the improvements brought by these terms.

\section{Conclusion}
In this work, we propose a computational efficient Monocular-to-3D Virtual Try-On Network (M3D-VTON) that builds on the merits of both 2D and 3D approaches to produce the 3D try-on mesh from 2D information. 
Our M3D-VTON decomposes the 3D try-on task into a 2D try-on and a body depth estimation problem. In future work, we will investigate if the two can further promote each other in a cyclic-manner. To get more realistic texture fusion results, M3D-VTON utilizes a two-stage warping strategy as well as segmentation and depth guidance .  We also introduce a novel depth gradient constraint to generate more detailed depth maps. Our method provides a faster and more economic solution for the monocular-to-3D virtual try-on task.  

\section{Acknowledgements}
This work was supported in part by National Key R\&D Program of China under Grant No. 2020AAA0109700, National Natural Science Foundation of China (NSFC) under Grant No.U19A2073 and No.61976233, Guangdong Province Basic and Applied Basic Research (Regional Joint Fund-Key) Grant No.2019B1515120039, Guangdong Outstanding Youth Fund (Grant No. 2021B1515020061), Shenzhen Fundamental Research Program (Project No. RCYX20200714114642083, No. JCYJ20190807154211365), Zhejiang Lab’s Open Fund (No. 2020AA3AB14).

{\small
\bibliographystyle{ieee_fullname}
\bibliography{egbib}
}

\end{document}